\documentclass[letterpaper]{article} 
\usepackage[]{aaai24}  
\usepackage{times}  
\usepackage{helvet}  
\usepackage{courier}  
\usepackage{amsmath}
\usepackage{amsfonts}
\usepackage{subfigure}
\usepackage{mathabx}
\usepackage{xcolor}
\usepackage{enumitem}
\usepackage{multirow}
\usepackage{booktabs}
\newcommand{\modi}{\textcolor{black}}
\usepackage[hyphens]{url}  
\usepackage{graphicx} 
\usepackage{natbib}  
\usepackage{caption} 
\usepackage{algorithm}
\usepackage{algorithmic}

\usepackage{newfloat}
\usepackage{listings}
\DeclareCaptionStyle{ruled}{labelfont=normalfont,labelsep=colon,strut=off} 
\floatstyle{ruled}
\newfloat{listing}{tb}{lst}{}
\floatname{listing}{Listing}
\title{Improving the Robustness of Knowledge-Grounded \\ Dialogue via Contrastive Learning}
\author{
    Jiaan Wang\textsuperscript{\rm 1}, Jianfeng Qu\textsuperscript{\rm 1}\thanks{Corresponding authors.}, Kexin Wang\textsuperscript{\rm 1}, Zhixu Li\textsuperscript{\rm 2}\footnotemark[1], Wen Hua\textsuperscript{\rm 3}, Ximing Li\textsuperscript{\rm 4}, An Liu\textsuperscript{\rm 1} 
}
\affiliations{
    \textsuperscript{\rm 1} School of Computer Science and Technology, Soochow University, Suzhou, China\\
    \textsuperscript{\rm 2} Shanghai Key Laboratory of Data Science, School of Computer Science, Fudan University, Shanghai, China\\
    \textsuperscript{\rm 3} The Hong Kong Polytechnic University, Hong Kong SAR, China\\
    \textsuperscript{\rm 4} College of Computer Science and Technology, Jilin University, China\\
    jawang.nlp@gmail.com, jfqu@suda.edu.cn, kxwang1@stu.suda.edu.cn, zhixuli@fudan.edu.cn

}

\usepackage{bibentry}

\begin{document}

\maketitle

\begin{abstract}
Knowledge-grounded dialogue (KGD) learns to generate an informative response based on a given dialogue context and external knowledge (\emph{e.g.}, knowledge graphs; KGs). Recently, the emergence of large language models (LLMs) and pre-training techniques has brought great success to knowledge-grounded dialogue. However, when building KGD systems in real applications, there are various real-world noises that are inevitable to face. For example, the dialogue context might involve perturbations such as misspellings and abbreviations. In addition, KGs typically suffer from incompletion and also might contain erroneous and outdated facts. Such real-world noises pose a challenge to the robustness of KGD systems and hinder their applications in the real world.
In this paper, we propose an entity-based contrastive learning framework for improving the robustness of KGD.
Specifically, we make use of the entity information in a KGD sample to create both its positive and negative samples which involve semantic-irrelevant and semantic-relevant perturbations, respectively.
The contrastive learning framework ensures the KGD model is aware of these two types of perturbations, thus generating informative responses with the potentially noisy inputs in real applications.
Experimental results on three benchmark datasets show that our method achieves new state-of-the-art performance in terms of automatic evaluation scores, verifying its effectiveness and potentiality.
Furthermore, we show that our method can generate better responses than comparison models in both the noisy and the few-shot settings.\footnote{\url{https://github.com/kxinwang2023/EnCo}}
\end{abstract}

\section{Introduction}

\begin{figure}[t]
\centerline{\includegraphics[width=0.40\textwidth]{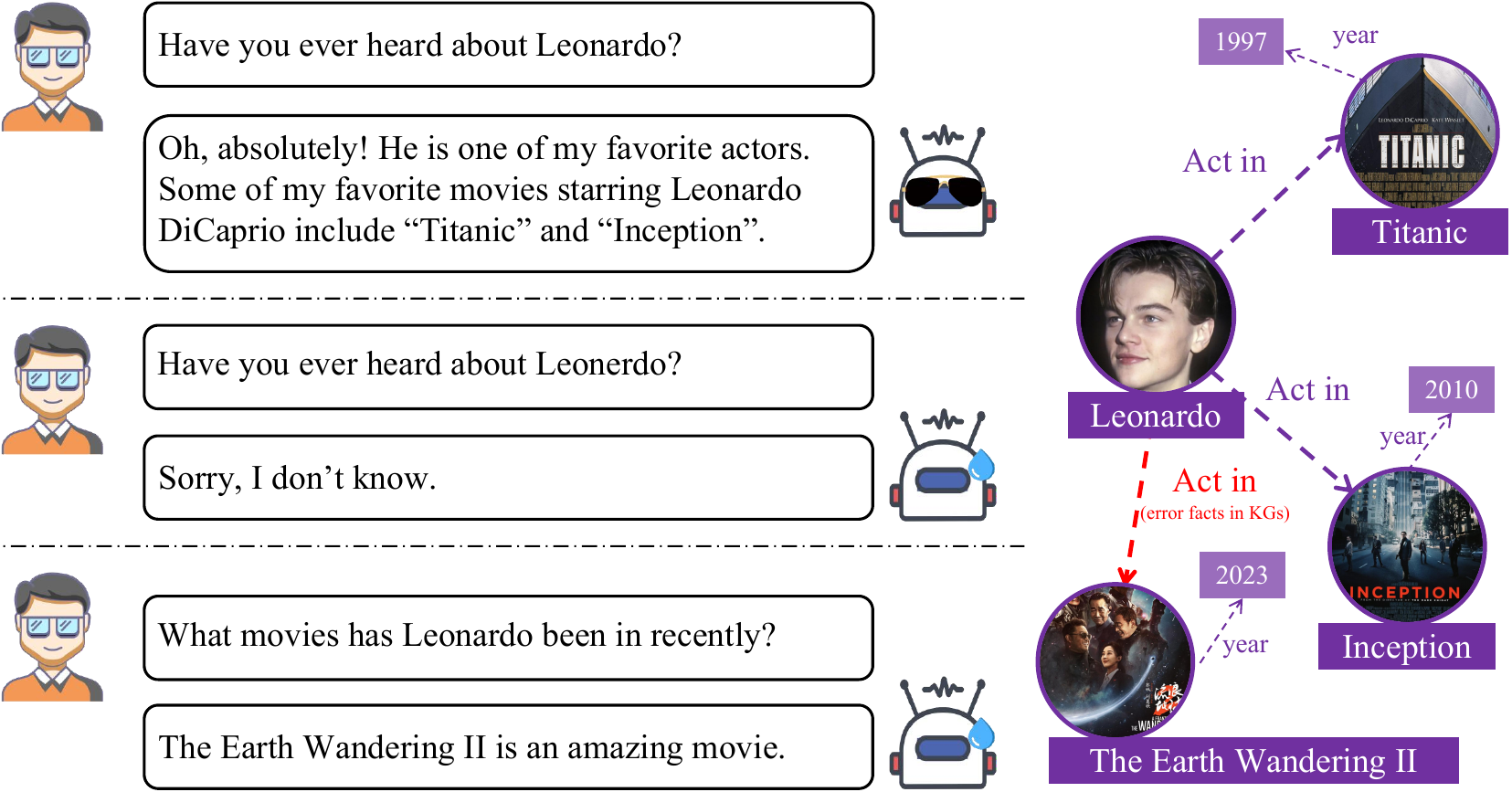}}
\caption{Illustrations of robustness in knowledge-grounded dialogue. The response in the first dialogue satisfies humans, while those in the second and the third dialogues do not due to misspellings and erroneous facts in the KG, respectively.}
\label{fig:intro}
\end{figure}

Knowledge-Grounded Dialogue (KGD) aims to generate an informative response based on a given dialogue context and external knowledge to improve the usefulness and meaningfulness of the generated responses~\cite{Ghazvininejad2017AKN,Zhou2018CommonsenseKA,Liu2019KnowledgeAC,Kim2020Sequential,Li2021KnowledgeGroundedDG,rashkin-etal-2021-increasing,wu2022section,sun-etal-2023-towards}.
As for the choice of knowledge source, structural knowledge graphs (KGs) are proven options~\cite{Moon2019OpenDialKGEC,Jung2020AttnIOKG,wu2022section}, which consist of a lot of knowledge facts that are frequently used in daily life~\cite{zheng2021enhancing,zhang2022aligning,cao2023pre,li2023attribute}.
Despite the great success that has been achieved by these efforts, especially along with the rapid development of large language models (LLMs), existing work neglects the robustness of KGD methods facing real-world noises.
As shown in Figure~\ref{fig:intro}, the dialogue context might involve inevitable perturbations like misspellings and abbreviations.
The KGs might also convey erroneous and outdated facts.
These perturbations pose a challenge to the robustness of KGD methods.

As large language models (LLMs) become a promising way toward artificial general intelligence, many dialogue systems utilize LLMs as their backbone to achieve strong conversational ability.
However, as pointed out by recent work~\cite{Wang2023OnTR}, there is still room for improving their robustness even for ChatGPT.
Besides, some studies~\cite{Zhu2023KTGATIT,wang2023chatgpt,wang-etal-2023-zero} show the robustness of knowledge-enhanced LLM is also limited.
Thus, improving the model robustness becomes one of the key challenges when deploying KGD models in real scenes.

In this paper, we propose an \textbf{en}tity-based \textbf{co}ntrastive learning (EnCo) framework for improving the robustness of KGD models.
Our key insight is to add perturbations in the given dialogue context as well as related knowledge, and let the KGD model learn how to generate an informative response with the potentially noisy inputs.
Specifically, for a vanilla KGD sample, we create its positive samples and negative samples, where the positive samples should share faithful (similar or clipped) semantics with the vanilla sample, and the negative samples involve conflict semantics with the vanilla ones. The object of EnCo ensures the model to represent similar KGD samples in a shared space by training the encoder to minimize the representation distance of them. For conflict samples, EnCo makes the encoder maximize their representations to let the model be aware of the perturbations in negative samples.
Ideally, once the KGD model has the ability to distinguish perturbations in the potentially noisy dialogue context or KGs, its robustness could be ensured in real applications.

Therefore, the key in our EnCo framework is how to create positive and negative samples that should reflect real-world noises as much as possible. (\romannumeral1) In positive samples, there might involve semantic-irrelevant perturbations in the real scene like misspellings and abbreviations (c.f., the second dialogue in Figure~\ref{fig:intro}). To this end, in view of paraphrasing which restates the same meaning in different lexical or syntactic expressions~\cite{bhagat2013paraphrase}, we utilize a paraphrasing model to create such positive samples whose dialogue contexts ideally share similar semantics with that in the vanilla sample.
However, in the preliminary experiments, we find that the paraphrasing model might change the entities in the vanilla context to other similar entities during paraphrasing. Though these paraphrased entities are relevant to the original ones, they also might introduce semantic-relevant perturbations to cause semantic gaps which should be avoided. Thus, we introduce a simple yet effective method (named entity-guided paraphrasing) to explicitly let the paraphrasing model be aware of the entities in the context and keep them unchanged during paraphrasing at both the training and inference stages.
(\romannumeral2) The negative samples should contain conflict semantics with the vanilla sample. In this manner, the KGD model can learn to distinguish the semantic-relevant perturbations in the real scene.
Existing contrastive learning models typically adopt the rest of the samples in the same mini-batch as negative samples~\cite{jaiswal2020survey,Poddar2022DialAugMU,jiang-etal-2023-vision} or retrieve negative samples from candidates~\cite{karpukhin-etal-2020-dense}, which do not meet our requirements.
To create such negative samples in our scene, we propose an entity-guided negative augmentation strategy that randomly deletes or replaces original entities in the vanilla context. To make the augmented negative samples hard to distinguish, we also edit the corresponding KG in the same way. For example, if we replace an entity $e_a$ in the vanilla context with another one $e_b$, we also replace $e_a$ with $e_b$ in the KG to eliminate the potential semantic gap between the context and the KG in the negative samples.
This is because the semantic gap might let the KGD model learn the shortcut from comparing the dialogue context and the KG (both of them are inputs to the model) to distinguish the semantic-relevant perturbations.

Experimental results on three KGD benchmark datasets show that our EnCo outperforms the previous state-of-the-art models in terms of widely-used automatic evaluation metrics, indicating its effectiveness and superiority.
We also create a robustness test set based on existing benchmark data and show that our method could improve the robustness of KGD models when faced with noisy inputs.
Furthermore, we also conduct few-shot evaluation and human evaluation to suggest that our method could generate satisfactory responses compared to existing related methods.

Our contributions are concluded as follows:
\begin{itemize}
\item To the best of our knowledge, we are the first to study the robustness of knowledge-grounded dialogue models, and we introduce an entity-based contrastive learning framework for improving the robustness of KGD models.
\item To let the KGD models be aware of semantic-relevant and sematic-irrelevant perturbations in real applications, we propose to utilize the entity information to guide the creation of positive and negative samples via paraphrasing and negative augmentation strategy, respectively.
\item Experimental results on three KGD benchmark datasets show that our method outperforms the previous state-of-the-art models. Robustness evaluation further demonstrates that our method could generate informative responses with noisy inputs. In addition, we show the few-shot ability of our method.
\end{itemize}

\begin{figure*}[t]
\centerline{\includegraphics[width=0.85\textwidth]{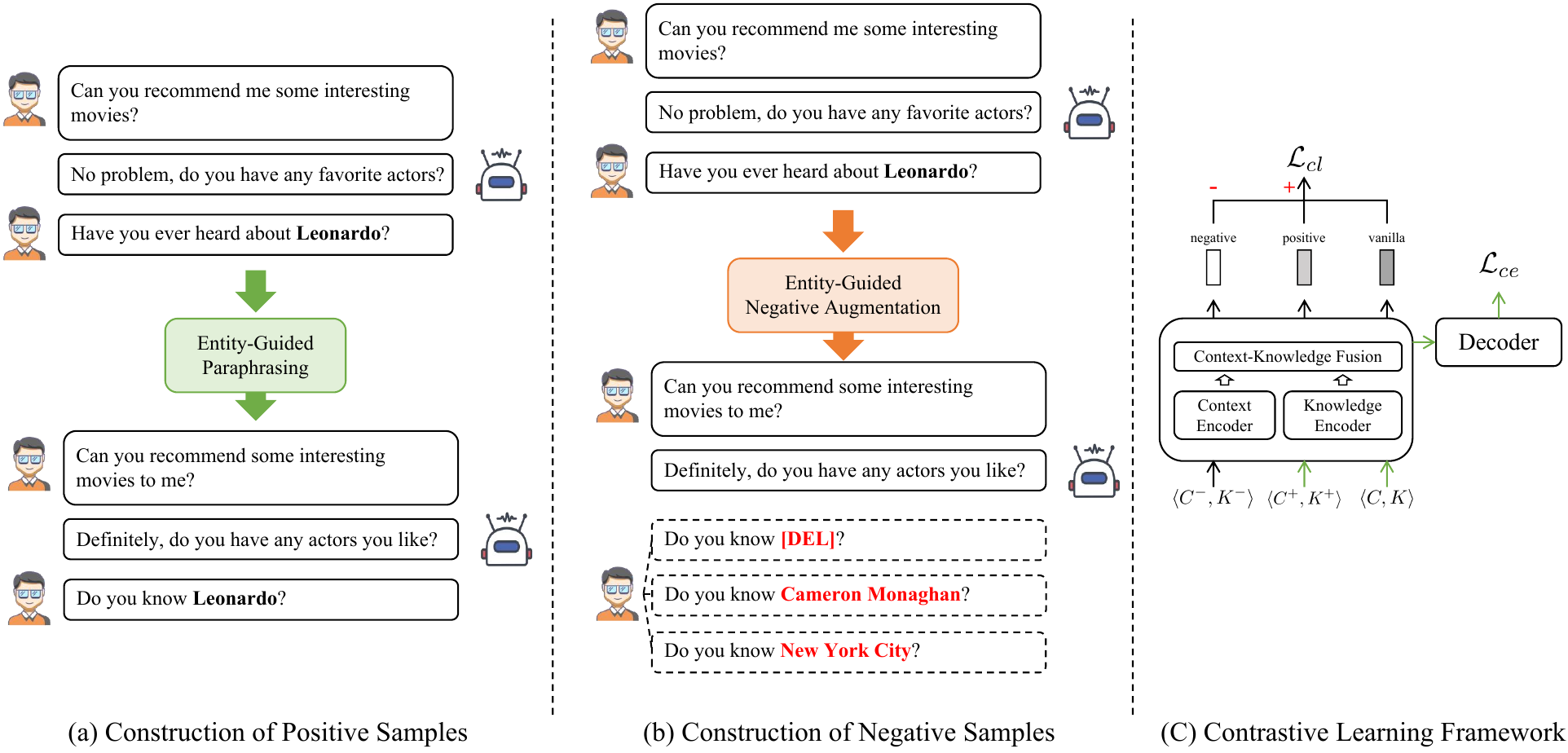}}
\caption{Illustrations of positive samples and negative samples in our entity-based contrastive learning framework.}
\label{fig:overview}
\end{figure*} 

\section{Related Work}
\subsection{Knowledge-Grounded Dialogue}

Knowledge-grounded dialogue (KGD) aims at incorporating external knowledge to generate informative responses.
Structural knowledge graphs (KGs) are proven options to serve as the knowledge source, providing a large number of knowledge facts in the real world~\cite{Moon2019OpenDialKGEC,Jung2020AttnIOKG,wu2022section,wang2022knowledge,wang2022incorporating,wang2023snowman,zhang2023aspectmmkg}. Many previous KGD work utilizes commonsense KGs~\cite{Zhou2018CommonsenseKA,Wu2020DiverseAI} or domain KGs~\cite{Wu2019ProactiveHC,Zhou2020KdConvAC,wang2022rt} to guide the dialogue generation.

Some researchers~\cite{vougiouklis2016neural,Ghazvininejad2017AKN} attempt to leverage memory networks~\cite{sukhbaatar2015end} to store the relevant knowledge and then generate responses conditioned on both the dialogue context and the stored knowledge.
To extract more relevant knowledge from the knowledge source, some work~\cite{Lian2019LearningTS,Wu2020DiverseAI} selects the knowledge by approximating the prior distribution, \emph{i.e.}, $P$(knowledge$|$context) based on the posterior distribution, \emph{i.e.}, $P$(knowledge$|$context, response),  leading to accurate knowledge selection and high-quality generated responses.
Besides, some researchers find that the response might involve words/tokens that existed in the knowledge resource (named knowledge words), and they employ various copy mechanisms to copy words or improve the generation probability of the knowledge words~\cite{lin-etal-2020-generating,Wu2020DiverseAI,bai2021learning,Liang2020InfusingMK}.
To effectively excavate the structural information of KGs, previous literature also utilizes graph neural networks (GNNs)~\cite{Liu2019KnowledgeAC,Moon2019OpenDialKGEC,Jung2020AttnIOKG,Wu2021KnowledgeAwareDG} or knowledge graph embeddings (KGE)~\cite{Zhou2018CommonsenseKA,wang2022rt} over KGs to obtain their structure-aware representation that is further incorporated into the dialogue generation process.
Different from the previous KGD work which typically focuses on how to select relevant knowledge and how to generate high-quality responses conditioned on the relevant knowledge, we are the first to study the robustness of KGD models when faced with real-world noises.

\subsection{Robustness of Dialogue Systems}

There is a few work studies the robustness of dialogue systems.
\citet{sengupta2021robustness} focus on the robustness of intent classification (IC) and slot labeling (SL) in
task-oriented dialog systems. They show that common noise types (such as misspellings) substantially degrade the accuracy of IC and SL sub-models.
\citet{Poddar2022DialAugMU} study the robustness of retrieval-based dialogue systems when faced with perturbations. They utilize contrastive learning as an auxiliary objective to learn robust dialogue context representations and make retrieval-based dialogue systems retrieve proper responses from candidates under the perturbation setting like truncation, word deletion and word reordering.
\citet{chen-etal-2023-towards-robust} find that the input orders of persona sentences significantly impact the quality and consistency of the personalized dialogue systems. They propose to learn robust representation under different persona orders and improve the consistency of response generation.

Existing robustness studies of dialogue systems generally focus on task-oriented dialogue, retrieval-based dialogue as well as personalized dialogue, whose settings are different from that of knowledge-grounded dialogue in our work.

\section{Methodology}

In this section, we first give the definition of the knowledge-ground dialogue (KGD) task and then elaborate on the details of our entity-based contrastive learning (EnCo) framework.
As illustrated in Figure~\ref{fig:overview}, for a given KGD sample, the EnCo framework utilizes an entity-guided paraphrasing model and entity-guided negative augmentation to construct positive and negative samples, respectively.
To encode the KGD samples, EnCo involves a context encoder to encode the dialogue context, and a knowledge encoder to encode the relevant knowledge triples extracted from KGs. Next, a context-knowledge fusion module is used to fuse the information of both dialogue context and external knowledge. Further, a decoder is employed to generate responses conditioned on the fused representation.
As for contrastive learning, EnCo makes the encoder minimize the representation distance between the given sample and the positive sample, and maximize that between the given sample and the negative sample.
In this manner, the KGD model can learn to distinguish both the semantic-relevant and the semantic-irrelevant perturbations, improving its robustness when faced with real-world noises.

\subsection{Task Definition}
\label{subsec:3.1}

Given a dialogue context $C = \{ u_1, u_2, ..., u_{n-1} \}$ and the corresponding relevant knowledge triple set
$K = \{(h_1,r_1,t_1), (h_2,r_2,t_2),...,(h_m, r_m, t_m)\}$, 
where $u_i$ represents the $i$-th utterance in the dialogue, $(h_j,r_j,t_j)$ denotes the head entity $h_j$ and tail entity $t_j$ have the relation $r_j$.
The goal of knowledge-grounded dialogue systems is to generate a proper response $u_n$ based on the dialogue context $C$ and the knowledge triples $K$.

\subsection{The Construction of Positive Samples}
\label{subsec:3.2}

To make the KGD model be aware of the semantic-irrelevant perturbations in the context $C$, we paraphrase the context to construct the positive samples for a given KGD sample. In this way, the paraphrased samples ideally share similar semantics with the vanilla ones but with different lexical or syntactic expressions.

Considering the changing of entities during paraphrasing might also change the semantics involved in the context, we design a simple yet effective entity-guided paraphrasing that explicitly models the entity information in the following two steps: (1) we first use an off-the-shelf named entity recognition (NER) toolkit (\emph{i.e.}, TexSmart\footnote{\url{https://texsmart.qq.com/en}}) to mine all the entities involved in the context $C$. Then, (2) we give the boundary information of entities by adding two special tokens: \texttt{[Ent]} and \texttt{[\textbackslash Ent]}. For example, when paraphrasing the sentence ``Do you know Leonardo?'', the NER toolkit first extracts the entity ``Leonardo'', and then the special tokens are added to the entity boundary to form the input of the paraphrase model: ``Do you know \texttt{[Ent]}Leonardo\texttt{[\textbackslash Ent]}?''. Thus, the paraphrase model could be aware of the entity information lying in the input sentences, and not change them during paraphrasing.

\vspace{0.5ex}
\noindent \textbf{The architecture of the paraphrase model.} We use BART-large model~\cite{Lewis2019BARTDS} as the backbone of the paraphrase model. This model contains standard transformer encoder-decoder architecture~\cite{Vaswani2017AttentionIA} (12 encoder layers, 12 decoder layers, 16 multi-head attention as well as 1024 hidden states). The BART model has been pre-trained on large-scale corpora with the self-supervised auto-denoising pre-training objectives~\cite{Lewis2019BARTDS}.

\vspace{0.5ex}
\noindent \textbf{Training of the paraphrase model.} ParaZh-22M~\cite{Hao2022ParaZh22MAL} is a large-scale paraphrase dataset with about 22M sentence pairs. We utilize the NER toolkit (TexSmart) to preprocess the dataset. In detail, we first extract the entities from each sentence pair, and then add the special tokens to the entity boundaries. To let the paraphrase model not change the entities involved in the input sentences, we only reserve the sentence pairs whose source and target sentences contain the same entity set, resulting in about 6.5M training samples. The paraphrase model is trained to generate the golden paraphrased sentences in the text-generation style:
\begin{equation}
p_{\theta}(\hat{s}|s) = \sum^{|\hat{s}|}_{t=1} p_{\theta} (\hat{s}_{t} | s, \hat{s}_{1:t-1})
\end{equation}
where $\theta$ denotes the parameters of the paraphrase model, $s$ and $\hat{s}$ indicate the input sentence and the golden paraphrased sentence, respectively. $\hat{s}_{1:t-1}$ is the partial paraphrasing.

\vspace{0.5ex}
\noindent \textbf{Inference of the paraphrase model.} After training the paraphrase model, the model is used to paraphrase the dialogue context $C = \{u_1, u_2, ..., u_{n-1}\}$. We input each utterance $u_i$ to the paraphrase model, and use the top-$k$ sampling strategy to decode the paraphrased utterance $\hat{u}_i$. All paraphrased utterances construct the context of the positive sample, which we denote as $C^{+} = \{ \hat{u}_1, \hat{u}_2, ..., \hat{u}_{n-1} \}$.

Furthermore, to improve the robustness of the KGD model when faced with incomplete knowledge, the knowledge triple set used in the positive samples (denoted as $K^{+}$) is truncated from the vanilla ones ($K$).
Specifically, we randomly remove $\lambda$\% triples in $K$ to obtain $K^{+}$, and $\lambda$ is randomly chosen from 0 to 15 for each sample.
Note that removing information does not introduce conflict semantics, thus the truncated knowledge set can be used in the positive samples, making the KGD model learn to still generate informative responses with incomplete grounding knowledge.

\subsection{The Construction of Negative Samples}
\label{subsec:3.3}

Different from the positive sample $\langle C^{+}, K^{+} \rangle$, a negative sample, denoted as $\langle C^{-}, K^{-} \rangle$, should introduce semantic-relevant perturbations to the given KGD sample $\langle C, K \rangle$.
In view of existing contrastive learning models, they could be classified into two methods: (1) adopting the rest of the samples in the same mini-batch as negative samples~\cite{jaiswal2020survey,jiang-etal-2023-vision} or (2) retrieving negative samples from candidates~\cite{karpukhin-etal-2020-dense}. In KGD, the rest samples in the same mini-batch only contain irrelevant information and do not introduce semantic-relevant perturbations. Besides, we do not have negative candidates in the KGD scene.
Thus, simply utilizing existing methods can not let the KGD model be aware of the semantic-relevant perturbations, making us decide to use the entity information lying in the context $C$ to construct negative samples.

We first make use of the NER toolkit (TextSmart) to mine all the entities involved in $C$, and for each entity, it has a 30\% probability to make one of the following changes (the remaining 70\% probability does not change anything):

\begin{itemize}[leftmargin=*,topsep=0pt]
\setlength{\itemsep}{0pt}
\setlength{\parsep}{0pt}
\setlength{\parskip}{0pt}
\item \emph{Randomly Deletion}: The entity will be removed from $C$, and we replace it with a special token ``\texttt{[DEL]}''.
\item \emph{Relevant Replacement}: The entity will be replaced with another entity that has the same entity type as the entity.
\item \emph{Irrelevant Replacement}: The entity will be replaced with another one that has a different type from the entity.
\end{itemize}

The above three changes share 10\%, 80\% and 10\% probabilities to employ for each of the changing entities (which are first selected with the 30\% probability).
In this manner, the changed context $C^{-}$ might involve the different degrees of semantic-relevant perturbations.
After that, the entities in the knowledge set $K$ also conduct the same changes as the context $C$ to obtain $K^{-}$. For example, if we delete an entity in $C$ and mark it as ``\texttt{[DEL]}'', the knowledge set $K$ will also delete it and denote it as ``\texttt{[DEL]}'' (if has).

\subsection{Contrastive Learning Framework}
\label{subsec:3.4}

After constructing the positive and negative samples, we use a KGD model to encode all the samples and minimize the representation distance between the vanilla sample and the positive sample while maximizing that between the vanilla sample and the negative sample.

\vspace{0.5ex}
\noindent \textbf{KGD model.} As shown in Figure~\ref{fig:overview} (c), the KGD model receives dialogue context $C$ and knowledge triple set $K$ as inputs and generates responses. The model contains the following four modules: (1) a \emph{context encoder} to calculate the context representation, which involves $N_e$ stacked transformer encoder layers, where each layer consists of two sub-layers, a multi-head self-attention sublayer (SelfAttn) and a position-wise feed-forward network (FFN) sub-layer:
\begin{equation}
\small
S^{\ell}_{C} = \operatorname{SelfAttn}(H^{\ell-1}_{C}) + H^{\ell-1}_{C}, S^{\ell}_{C} \in \mathbb{R}^{d}
\end{equation}
\begin{equation}
\small
H^{\ell}_{C} = \operatorname{FFN}(S^{\ell}_{C}) + S^{\ell}_{C}, H^{\ell}_{C} \in \mathbb{R}^{d}
\end{equation}
where $H^{\ell-1}_{C}$ and $H^{\ell}_{C}$ denote the inputs and outputs of the $\ell$-th layer, respectively, and $H^0_C$ is initialized as the embedding of input context $C$ and $d$ is the hidden dimension.

(2) A \emph{knowledge encoder} is employed to calculate the knowledge representation. Following~\citet{wang2022rt}, we use TransR~\cite{Lin2015LearningEA} to obtain the representation of each knowledge triple $(h_i, r_i, t_i) \in K$:
\begin{equation}
\small
e_{h_i}, e_{r_i}, e_{t_i} \gets \operatorname{TransR}(h_i), \operatorname{TransR}(r_i), \operatorname{TransR}(t_i)
\end{equation}
\begin{equation}
\small
h_{k_i} = e_{h_i} \oplus e_{r_i} \oplus e_{t_i}
\end{equation}
where $e_{h_i}$, $e_{r_i}$ and $e_{t_i}$ indicate the representations of $h_i$, $r_i$ and $t_i$, respectively. $h_{k_i}$ denotes the representation of the knowledge triple $k_i = (h_i, r_i, t_i)$ and $\oplus$ means concatenation. In this way, we obtain all triple representations $\{h_{k_1}, h_{k_2}, ..., h_{k_m}\}$ via the knowledge encoder.

(3) a \emph{context-knowledge fusion module} to fuse the context and the knowledge representations. We adopt multi-head attention to fuse each triple representation $h_{k_i}$ with the context information:
\begin{equation}
\small
h^c_{k_i} = \operatorname{\parallel}\limits_{h=1}^{m} \operatorname{Attn}_h (h_{k_i} \gets H^{N_e}_C)
\end{equation}
\begin{equation}
\small
\operatorname{Attn}_h (h_{k_i} \gets H^{N_e}_C) = \sum_{j} \operatorname{softmax}(\frac{Q_h(h_{k_i}) \cdot K_h(H^{N_e}_{j})}{\sqrt{d}}) \cdot V_h(H^{N_e}_{j})
\end{equation}
where $h^{c}_{k_i}$ indicates the context-enriched representation of triple $k_i$. $\operatorname{\parallel}\limits_{h=1}^{m}$ indicates multi-head attention and $m$ is the number of multi-head attention. $\operatorname{Attn}_h$ denotes the $h$-th head, whose query vector, key vector and value vector are denoted as $Q_h(\cdot)$, $K_h(\cdot)$ and $V_h(\cdot)$, respectively. $H^{N_e}_{j}$ means the $j$-th vector in $H^{N_e}_C$ that also indicates the representation of the $j$-th token in the context $C$.

(4) A \emph{decoder} is used to generate responses, which consists of $N_d$ stacked transformer decoder layers. To let the decoder be conditioned on both the context and the knowledge, we modified the vanilla transformer decoder layer by adding parallel cross-attention over context information $H^{N_e}_C$ and knowledge information $H^c_K = [ h^c_{k_1}; h^c_{k_2}; ...; h^c_{k_m}]$:
\begin{equation}
\small
S^\ell_{dec} = \operatorname{SelfAttn}(H^{\ell-1}_{dec}) + H^{\ell-1}_{dec}
\end{equation}
\begin{equation}
\small
C^{\ell}_{dec} = \operatorname{CrossAttn}(S^\ell_{dec}, H^{N_e}_C) + \operatorname{CrossAttn}(S^\ell_{dec}, H^c_K) + S^\ell_{dec}
\end{equation}
\begin{equation}
\small
H^{\ell}_{dec} = \operatorname{FFN}(C^{\ell}_{dec}) + C^{\ell}_{dec}
\end{equation}
where $H^{\ell}_{dec}$ denotes the state of the $\ell$-th decoder layer. Then, at each decoding time step $t$, the top-layer ($N_d$-th) decoder hidden state $H^{N_d}_{dec,t}$ is fed into the softmax layer to produce the probability distribution of the next target token as:
\begin{equation}
\small
p(u^{t}_{n}| C,K, u^{<t}_{n}) = \operatorname{Softmax}(W_o H^{N_d}_{dec,t} + b_o)
\end{equation}
where $W_o$ and $b_o$ are trainable parameters, and $u^{t}_{n}$ denote the $t$-th token in the golden response $u_n$.

We use the vanilla sample and the positive samples to calculate cross-entropy loss:
\begin{equation}
\small
\mathcal{L}_{van} = - \sum^{|u_n|}_{t=1}\operatorname{log}(p(u^{t}_{n}| C,K, u^{<t}_{n}))
\end{equation}
\begin{equation}
\small
\mathcal{L}_{pos} = - \sum_{(C^{+}, K^{+}) \in S^{pos}} \sum^{|u_n|}_{t=1}\operatorname{log}(p(u^{t}_{n}| C^{+},K^{+}, u^{<t}_{n}))
\end{equation}
\begin{equation}\label{eq:ce_loss}
\small
\mathcal{L}_{ce} = \mathcal{L}_{van} + \mathcal{L}_{pos}
\end{equation}
where $S^{pos}$ denotes the set of positive samples.

\vspace{0.5ex}
\noindent \textbf{Contrastive Loss.} We also adopt the following contrastive loss over the vanilla, positive and negative KGD samples:
\begin{equation}
\small
\mathcal{L}_{ctr} = - \sum_{C^{+}} \sum_{C^{-}} \operatorname{log} \frac{f(H^{N_e}_C, H^{N_e}_{C^+})}{f(H^{N_e}_C, H^{N_e}_{C^+})+f(H^{N_e}_C, H^{N_e}_{C^-})}
\end{equation}
where $f(a,b) = exp(a^{\top}b)$. Therefore, our final loss is:
\begin{equation}\label{eq:ctr_loss}
\mathcal{L} = \mathcal{L}_{ce} + \alpha \mathcal{L}_{ctr}
\end{equation}

\section{Experiments}

\subsection{Implementation Details}
We implement our EnCo framework with PyTorch and Huggingface Transformers\footnote{\url{https://github.com/huggingface/transformers}}~\cite{wolf2019huggingface} libraries. The context encoder and decoder are initialized by a pre-trained BART-large model\footnote{\url{https://huggingface.co/fnlp/bart-large-chinese}}, \emph{i.e.}, the numbers ($N_e$ and $N_d$) of encoder layers as well as decoder layers are 12, and the hidden dimension $d$ is 1,024. The number of multi-head attention in the context-knowledge fusion module ($m$) is set to 8. Following~\citet{wang2022rt}, the embedding size of entities and relations is set to 200, and the implementation of TransR is based on the OpenKE toolkit\footnote{\url{https://github.com/thunlp/OpenKE}}.
For each KDG sample, we create 5 positive and 5 negative samples.
We train the KGD model on two 32GB Tesla V100 GPU. We set the hyperparameters based on the preliminary experiments on the development set. We leverage the Adam optimizer with a default initial momentum and adopt linear warmup in the first 1,000 steps. The mini-batch size is set to 8, and the coefficient $\alpha$ in the final loss function is set to 1.0. We use minimal hyperparameter tuning using Learning Rates (LRs) in [1e5, 2e-5, 3e-5, 5e-5] and epochs of 10 to 20. We find the model with LR of 5e-5 and 20 epochs to work best.
All experimental results listed in this paper are the average of 3 runs.

\begin{table*}[t]
\centering
\resizebox{0.98\textwidth}{!}
{
\begin{tabular}{lccccccccccccccccccccccccccc}
\toprule[1pt]
\multicolumn{1}{c}{\multirow{2}{*}{Model}} & \multicolumn{9}{c}{KdConv (music)}                                                                                                                   & \multicolumn{9}{c}{KdConv (travel)}                                                                                                                  & \multicolumn{9}{c}{KdConv (film)}                                                                                                                    \\
\cmidrule(r){2-10} \cmidrule(r){11-19} \cmidrule(r){20-28} \multicolumn{1}{c}{}                       & PPL ($\downarrow$)           & \multicolumn{4}{c}{BLEU-1/2/3/4 ($\uparrow$)}                                  & \multicolumn{4}{c}{DISTINC-1/2/3/4 ($\uparrow$)}                              & PPL ($\downarrow$)          & \multicolumn{4}{c}{BLEU-1/2/3/4 ($\uparrow$)}                                  & \multicolumn{4}{c}{DISTINC-1/2/3/4 ($\uparrow$)}                              & PPL ($\downarrow$)           & \multicolumn{4}{c}{BLEU-1/2/3/4 ($\uparrow$)}                                  & \multicolumn{4}{c}{DISTINC-1/2/3/4 ($\uparrow$)}                              \\ \midrule[1pt]
Seq2Seq                                    & 16.17         & 28.89          & 16.56          & 10.63          & 7.13           & 2.52          & 7.02           & 12.69          & 18.78          & 10.44         & 29.61          & 20.04          & 14.91          & 11.74          & 3.75          & 11.15          & 19.01          & 27.16          & 23.88         & 26.97          & 14.31          & 8.53           & 5.30           & 2.51          & 7.14           & 13.62          & 21.02          \\
HRED                                       & 16.82         & 29.92          & 17.31          & 11.17          & 7.52           & 2.71          & 7.71           & 14.07          & 20.97          & 10.90         & 30.92          & 20.97          & 15.61          & 12.30          & 4.15          & 12.01          & 20.52          & 28.74          & 24.74         & 27.03          & 14.07          & 8.30           & 5.07           & 2.55          & 7.35           & 14.12          & 21.86          \\
Seq2Seq+Know.                              & 17.12         & 29.60          & 17.26          & 11.36          & 7.84           & 3.93          & 12.35          & 23.01          & 34.23          & 10.62         & 37.04          & 27.28          & 22.16          & 18.94          & \textbf{4.25} & 13.64          & 24.18          & 34.08          & 25.56         & 27.45          & 14.51          & 8.66           & 5.32           & 2.85          & 7.98           & 15.09          & 23.17          \\
HRED+Know.                                 & 17.69         & 29.73          & 17.51          & 11.59          & 8.04           & 3.80          & 11.70          & 22.00          & 33.37          & 11.15         & 36.87          & 26.68          & 21.31          & 17.96          & 3.98          & 13.31          & 24.06          & 34.35          & 26.27         & 27.94          & 14.69          & 8.73           & 5.40           & 2.86          & 8.08           & 15.81          & 24.93          \\
KIC                                        & 13.06         & 30.41          & 18.48          & 13.87          & 9.42           & 3.31          & 12.77          & 23.49          & 33.91          & 8.46          & 37.21          & 28.89          & 23.30          & 19.94          & 3.45          & 13.74          & 25.47          & 35.01          & 11.29         & 28.12          & 15.17          & 9.53           & 7.20           & 2.63          & 14.38          & 26.74          & 38.49          \\
SDAN                                       & 14.78         & 30.92          & 18.92          & 14.40          & 10.54          & 4.03          & 12.61          & 23.07          & 32.71          & 9.32          & 38.13          & 30.49          & 24.96          & 21.16          & 3.62          & 13.86          & 25.31          & 34.83          & 14.52         & 28.96          & 16.72          & 10.22          & 7.91           & 3.25          & 11.56          & 23.47          & 33.81          \\
\modi{KSPN}   &  \modi{3.59}         & \modi{35.55}         & \modi{26.90}        & \modi{23.77}    & \modi{18.37}        & \modi{3.32}        & \modi{15.93}       & \modi{29.22}    & \modi{40.16}        &   \modi{2.51}    &  \modi{44.39}   &  \modi{38.10}    &  \modi{33.85}   &  \modi{31.71}        &  \modi{2.89}       &  \modi{15.42}   &  \modi{25.96}   &  \modi{34.51}      & \modi{3.80}    & \modi{29.44}       &\modi{19.12}  & \modi{14.92}         & \modi{12.01}          & \modi{3.03}         & \modi{16.04}        & \modi{30.27}       &  \modi{43.66}         \\
BART                                       & \textbf{2.44} & 32.27          & 23.40          & 18.44          & 15.22          & 2.80          & 13.68          & 25.19          & 35.61          & \textbf{1.69} & 36.61          & 30.29          & 26.54          & 23.92          & 2.56          & 13.58          & 22.85          & 30.87          & \textbf{2.82} & 29.68          & 20.43          & 15.26          & 11.97          & 2.50 & 15.12          & 27.96          & 39.56          \\
BART+Know                                  & 2.89          & 34.21          & 26.14          & 23.11          & 17.98          & 3.04          & 15.79          & 28.09          & 38.46          & 1.88          & 39.45          & 33.72          & 30.55          & 28.78          & 3.10          & 15.05          & 24.80          & 33.41          & 3.06          & 29.98          & 20.68          & 15.44          & 12.21          & 2.89          & 15.97          & 29.02          & 42.73          \\ 
\modi{KRP-DS}   & \modi{3.05}         & \modi{36.88}          & \modi{27.71}         & \modi{23.89}     & \modi{18.92}          & \modi{3.55}         & \modi{16.11}          & \modi{29.47}      & \modi{40.78}         &   \modi{2.08}       &  \modi{45.00}        &   \modi{38.52}     &    \modi{34.69}      &   \modi{32.09}        &    \modi{2.71}       &   \modi{15.82}       &   \modi{26.59}     &    \modi{35.31}      & \modi{3.17}      & \modi{30.14}        & \modi{20.79}    & \modi{15.80}          & \modi{12.74}           & \modi{3.29}         & \modi{16.12}        &  \modi{30.84}       & \modi{43.90}         \\ \midrule[1pt]
EnCo (Our)                                      & 3.91          & \textbf{39.39} & \textbf{30.18} & \textbf{25.11} & \textbf{20.81} & \textbf{4.23} & \textbf{18.05} & \textbf{31.04} & \textbf{44.47} & 2.57          & \textbf{46.61} & \textbf{40.58} & \textbf{37.02} & \textbf{34.14} & 3.82          & \textbf{16.43} & \textbf{28.74} & \textbf{37.15} & 3.53          & \textbf{31.43} & \textbf{21.77} & \textbf{16.48} & \textbf{13.17} & \textbf{3.65}          & \textbf{17.43} & \textbf{33.29} & \textbf{47.36} \\ \bottomrule[1pt]
\end{tabular}
}
\caption{Experimental results on the KdConv dataset. The \textbf{bold} denotes the best performance. $\uparrow$ indicates higher is better. $\downarrow$ indicates lower is better.}
\label{table:kdconv_results}
\end{table*}

\begin{table}[t]
\centering
\resizebox{0.48\textwidth}{!}
{
\begin{tabular}{lccccccccc}
\toprule[1pt]
              & PPL ($\downarrow$)           & \multicolumn{4}{c}{BLEU-1/2/3/4 ($\uparrow$)}                                  & \multicolumn{4}{c}{DISTINC-1/2/3/4 ($\uparrow$)}                              \\ \midrule[1pt]
Seq2Seq       & 9.14          & 26.49          & 16.97          & 12.55          & 8.81           & 3.16          & 7.49           & 15.02          & 24.13          \\
HRED          & 22.82         & 32.71          & 18.16          & 14.89          & 9.70           & 4.11          & 8.98           & 16.21          & 24.95          \\
Seq2Seq+Know. & 10.96         & 28.32          & 18.61          & 13.96          & 9.32           & 4.30          & 9.20           & 16.48          & 25.67          \\
HRED+Know.    & 24.30         & 34.67          & 19.81          & 15.11          & 11.67          & 4.71          & 10.57          & 17.68          & 26.46          \\
KIC           & 10.36         & 37.70          & 26.22          & 22.09          & 17.92          & 6.31          & 24.78          & 37.66          & 48.91          \\
SDAN          & 8.42          & 36.47          & 20.46          & 17.31          & 12.91          & 5.53          & 12.39          & 18.60          & 28.01          \\
MGCG\_G       & 6.43          & 35.14          & 22.91          & 19.10          & 16.23          & 6.17          & 19.41          & 28.81          & 31.76          \\
\modi{KSPN} & \modi{4.02} & \modi{38.92} & \modi{24.51} & \modi{17.01} & \modi{13.75} & \modi{6.02} & \modi{24.98} & \modi{37.02} & \modi{47.55} \\ 
BART          & \textbf{3.20} & 36.54          & 25.17          & 20.07          & 18.86          & 5.33          & 23.12          & 35.75          & 46.82          \\
BART+Know     & 3.86          & 38.91          & 28.50          & 23.71          & 20.10          & 5.89          & 24.49          & 36.39          & 47.31          \\ 
\modi{KRP-DS}   & \modi{5.06}         & \modi{39.06}          & \modi{28.64}         & \modi{23.96}     & \modi{20.66}          & \modi{5.83}         & \modi{24.02}          & \modi{36.91}      & \modi{47.82}  \\ \midrule[1pt]
EnCo (Our)         & 5.57          & \textbf{40.10} & \textbf{29.69} & \textbf{24.89} & \textbf{21.30} & \textbf{6.88} & \textbf{25.66} & \textbf{38.50} & \textbf{49.17} \\ \bottomrule[1pt]
\end{tabular}
}
\caption{Experimental results on the DuConv dataset.}
\label{table:duconv_results}
\end{table}

\begin{table}[t]
\centering
\resizebox{0.48\textwidth}{!}
{
\begin{tabular}{lccccccccc}
\toprule[1pt]
              & PPL ($\downarrow$)          & \multicolumn{4}{c}{BLEU-1/2/3/4 ($\uparrow$)}                                  & \multicolumn{4}{c}{DISTINC-1/2/3/4 ($\uparrow$)}                              \\ \midrule[1pt]
Seq2Seq       & 20.10         & 16.19          & 9.31           & 5.41           & 2.47           & 0.42          & 1.02           & 8.45           & 15.35          \\
HRED          & 21.56         & 22.76          & 15.71          & 6.84           & 4.13           & 0.72          & 2.56           & 10.81          & 17.66          \\
Seq2Seq+Know. & 22.82         & 18.81          & 10.20          & 6.34           & 3.03           & 0.66          & 1.34           & 9.92           & 16.22          \\
HRED+Know.    & 23.96         & 24.33          & 16.15          & 8.94           & 5.77           & 1.03          & 3.97           & 12.24          & 19.72          \\
KIC           & 15.59         & 31.78          & 21.56          & 17.73          & 14.52          & 1.52          & 10.91          & 19.67          & 28.66          \\
SDAN          & 19.72         & 26.11          & 19.03          & 10.17          & 8.69           & 2.17          & 5.32           & 14.01          & 21.92          \\
MGCG\_G       & 14.89         & 36.22          & 25.23          & 21.79          & 17.05          & 2.33          & 8.16           & 15.26          & 23.02          \\
\modi{KSPN}   & \modi{5.50}         & \modi{38.01}          & \modi{28.31}         & \modi{23.19}     & \modi{19.80}          & \modi{2.13}         & \modi{12.91}          & \modi{18.67}      & \modi{30.70}  \\
BART          & \textbf{2.92} & 37.95          & 28.12          & 22.70          & 18.96          & 2.41          & 13.42          & 20.83          & 31.57          \\
BART+Know     & 3.42          & 39.17          & 28.99          & 24.32          & 20.82          & 2.63          & 16.35          & 24.67          & 36.21          \\ 
\modi{KRP-DS}   & \modi{3.50}         & \modi{39.02}          & \modi{29.12}         & \modi{24.41}     & \modi{20.95}          & \modi{2.51}         & \modi{16.62}          & \modi{25.03}      & \modi{37.29} \\ \midrule[1pt]
EnCo (Our)         & 4.36          & \textbf{40.05} & \textbf{29.55} & \textbf{24.68} & \textbf{21.26} & \textbf{2.78} & \textbf{17.49} & \textbf{31.48} & \textbf{43.88} \\ \bottomrule[1pt]
\end{tabular}
}
\caption{Experimental results on the DuRecDial dataset.}
\label{table:durecdial_results}
\end{table}

\subsection{Experimental Setups}

\noindent \textbf{Datasets.} We conduct experiments on three widely-used public KGD datasets: (1) KdConv~\cite{Zhou2020KdConvAC} involves 4.5K dialogues and 86K utterances from music, travel and film domains. There are a total of 85K KGD samples in KdConv. The knowledge in KdConv contains 13.1K entities, 9.1K relations and 157.0K triples.
(2) DuConv~\cite{Wu2019ProactiveHC} contains 180K samples about film and entertainment (29K dialogues as well as 270K utterances). The knowledge in DuConv involves about 3.6M triples.
(3) DuRecDial~\cite{Liu2020TowardsCR} has 145K samples with 10.2K dialogues. It also contains 21.8K entities, 454 relations and 222.2K triples.
All samples in these three datasets are annotated with dialogue-level relevant knowledge triples.

\vspace{0.5ex}
\noindent \textbf{Metrics.} Following previous work~\cite{Zhou2020KdConvAC,Liu2020TowardsCR}, we adopt perplexity (PPL), BLEU~\cite{papineni2002bleu} and DISTINCT~\cite{li2016diversity} to measure the fluency, relevance and diversity of the generated responses.

\vspace{0.5ex}
\noindent \textbf{Baselines.} We compare the proposed method with the following baselines:
(1) \textbf{Seq2Seq}~\cite{luong-etal-2015-effective} is a stack RNN-based model with attention mechanisms;
(2) \textbf{HRED}~\cite{Serban2015BuildingED} is a hierarchical RNN-based generative model;
(3) \textbf{Seq2Seq+Know} and (4) \textbf{HRED+Know} are introduced by~\citet{Zhou2020KdConvAC}, which fuse the context vector with the knowledge vector to form the initial state of the decoder;
(5) \textbf{KIC}~\cite{lin-etal-2020-generating} uses recurrent knowledge interaction among response decoding steps to incorporate appropriate knowledge in dialogues;
%
(6) \textbf{SDAN}~\cite{Cui2021SyntacticallyDA} utilizes a knowledge-aware network to improve informativeness, and a syntactic latent variable network to generate syntactically diverse responses;
(7) \textbf{MGCG\_G}~\cite{Liu2020TowardsCR} consists of a goal-planning module and a goal-guided responding module to handle multi-type dialogs;
%
(8) \textbf{KSPN}~\cite{liu2022improving} proposes a knowledge selection-guided pointer network into the decoder to generate responses with the words from the captured knowledge;
(9) \textbf{BART}~\cite{Lewis2019BARTDS} is a transformer-based generative model which has been pre-trained with auto-denoising objectives;
(10) \textbf{BART+Know} uses our knowledge encoder and context-knowledge fusion module to incorporate knowledge into BART;
(11) \textbf{KRP-DS}~\cite{he2023krp} utilizes contextual information for path reasoning and guides knowledge prediction in KGD.

\subsection{Main Results}

Table~\ref{table:kdconv_results}, Table~\ref{table:duconv_results} and Table~\ref{table:durecdial_results} show the results on KdConv, DuConv and DuRecDial, respectively.\footnote{We do not report the performance of MGCG\_G baseline on the KdConv dataset since MGCG\_G mainly models the multiple dialogue types in every single dialogue. However, this situation does not appear in KdConv.} As we can see, the results on the three datasets indicate a similar trend. Specifically, EnCo achieves the best performance in terms of most metrics. Compared with the second-performance baseline (\emph{i.e.}, \modi{KRP-DS}), EnCo is significantly better than it with t-test p $<$ 0.01 in terms of BLEU and DISTINC scores, showing its effectiveness.
The positive samples created in our EnCo framework could also be regarded as a data augmentation approach for KGD, making the KGD model learn from diverse samples.
In terms of PPL, BART achieves the best performance. Besides, we find that the PPL generally increases when incorporating knowledge into KGD models. For example, HRED achieves 16.82 PPL on KdConv (music) while the counterpart of HRED+Know is 17.69 (+0.87).
We conjecture this is because the knowledge will influence the model's output distributions during inferences, and thus increase PPL. This situation has also been found in previous work~\cite{Zhou2020KdConvAC,wang2022rt}.

\begin{table*}[t]
\centering
\resizebox{0.90\textwidth}{!}
{
\begin{tabular}{lccccccccc}
\toprule[1pt]
\multicolumn{1}{c}{\multirow{2}{*}{Model}} & Vanilla       & Word          & Word          & Synonym       & Utterance     & \multirow{2}{*}{Paraphrasing} & Noises  & \modi{Entity} & \modi{Entity}      \\
\multicolumn{1}{c}{}                       & Test          & Deletion      & Replacement   & Replacement   & Deletion      &                               & (ChatGPT) & \modi{Deletion} & \modi{Replacement}     \\ \midrule[1pt]
HRED+Know.                                 & 34.67 / 19.81 & 29.89 / 19.63 & 25.44 / 15.01 & 28.57 / 18.16 & 30.12 / 19.72 & 28.86 / 18.48                 & 26.78 / 17.03  & \modi{29.11 / 18.54}  &  \modi{27.52 / 17.30} \\
BART+Know                                  & 38.91 / 28.50 & 34.10 / 23.84 & 32.37 / 21.94 & 33.50 / 23.12 & 34.03 / 23.62 & 33.29 / 22.95                 & 31.58 / 19.94 & \modi{32.91 / 22.15}  &  \modi{31.77 / 21.09}  \\
EnCo                                       & \textbf{40.10} / \textbf{29.69} & \textbf{36.03} / \textbf{25.76} & \textbf{34.49} / \textbf{24.05} & \textbf{35.49} / \textbf{25.09} & \textbf{35.10} / \textbf{24.70} & \textbf{35.33} / \textbf{24.97}                 & \textbf{34.07} / \textbf{22.62} & \modi{\textbf{37.42} / \textbf{27.02}}  & \modi{\textbf{36.11} / \textbf{26.18}}   \\ \bottomrule[1pt]
\end{tabular}
}
\caption{Experimental results of robustness study in terms of BLEU-1 and BLEU-2.}
\label{table:robustness_results}
\end{table*}

\subsection{Robustness Study}

To test the model's performance when faced with real-world noises, we add the following noises into the test set of DuConv, respectively: (1) randomly deleting 30\% words in the dialogue context; (2) randomly replacing 30\% words in the context with other words from the whole vocabulary; (3) randomly replacing 30\% noun words with their synonyms\footnote{\url{https://github.com/chatopera/Synonyms}}; (4) randomly deleting 30\% utterances in the context except for the last utterance; (5) paraphrasing the dialogue context using our entity-guided paraphrase model; (6) introducing noises by ChatGPT with the prompt of ``Dialogues in the real world can be noisy. For example, the original dialogue: \texttt{[$D$]} in the real world might be like: \texttt{[$\hat{D}$]}. Please generate the noisy version of the following dialogue: \texttt{[$D_i$]}''. $D$ and $\hat{D}$ denote an original and noisy dialogue pair, which is written by our human annotators and serves as an in-context sample for ChatGPT to ground. There are a total of 20 in-context samples \modi{that involve different kinds of real-world noises like misspellings and abbreviations}, and we randomly select one sample to prompt ChatGPT at each time. $D_i$ is the input dialogue from the test set of DuConv. \modi{(7) randomly deleting 30\% entities in KG; (8) randomly replacing 30\% entities in KG with other entities.}

Table~\ref{table:robustness_results} lists the results of the robustness study. Our EnCo achieves the best performances among all noisy testing in terms of BLEU-1 and BLEU-2 (other metrics also show the same situations), verifying its effectiveness when faced with real-world noises. The positive and negative samples in EnCo let the KGD model be aware of both the semantic-irrelevant and the semantic-relevant perturbations, thus improving the model's robustness in real applications.

\begin{table}[t]
\centering
\resizebox{0.45\textwidth}{!}
{
\begin{tabular}{lcc}
\toprule[1pt]
          & KdConv (music)                & KdConv (travel)               \\ \midrule[1pt]
EnCo      & \textbf{39.39} / \textbf{30.18} / \textbf{25.11} / \textbf{20.81} & \textbf{46.61} / \textbf{40.58} / \textbf{37.02} / \textbf{34.14} \\
\quad w/o $\mathcal{L}_{pos}$  & 36.59 / 28.01 / 23.54 / 18.73 & 43.29 / 37.51 / 34.05 / 31.74 \\
\quad w/o $\mathcal{L}_{ctr}$  & 36.02 / 27.19 / 23.21 / 18.44 & 42.80 / 36.95 / 33.47 / 31.29 \\
BART+Know & 34.21 / 26.14 / 23.11 / 17.98 & 39.45 / 33.72 / 30.55 / 28.78 \\ \bottomrule[1pt]
\end{tabular}
}
\caption{Results of ablations (BLEU-1/2/3/4) on DuConv.}
\label{table:ablation_results}
\end{table}

\subsection{Ablation Results}
Compared with the BART+Know baseline, our EnCo adopts two improvements: (1) the positive samples are used as data augmentation (\emph{i.e.}, the $\mathcal{L}_{pos}$ in Eq.~\ref{eq:ce_loss}); (2) the contrastive learning framework (\emph{i.e.}, the $\mathcal{L}_{ctr}$ in Eq.~\ref{eq:ctr_loss}). We conduct ablation studies to investigate the effect of $\mathcal{L}_{pos}$ and $\mathcal{L}_{ctr}$ by removing each of them. As shown in Table~\ref{table:ablation_results}, when removing either $\mathcal{L}_{pos}$ or $\mathcal{L}_{ctr}$, the model performance will decreases.
Thus, the effectiveness of both improvements is proven.

\begin{figure}[t]
\centering
\subfigure{
  \includegraphics[width=0.47\linewidth]{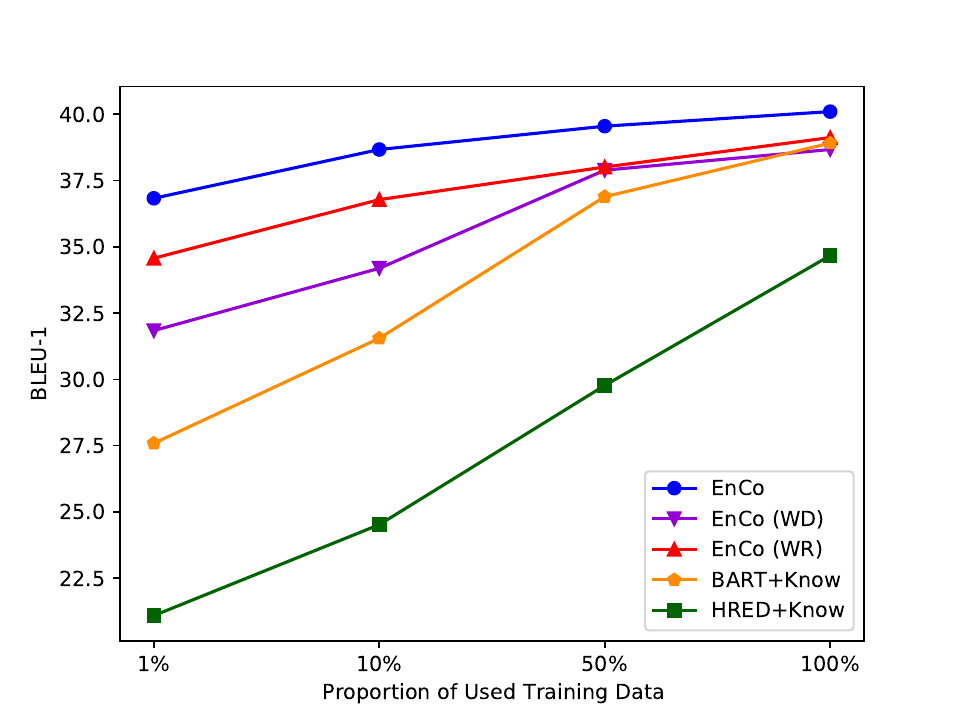}
}
\subfigure{
  \includegraphics[width=0.46\linewidth]{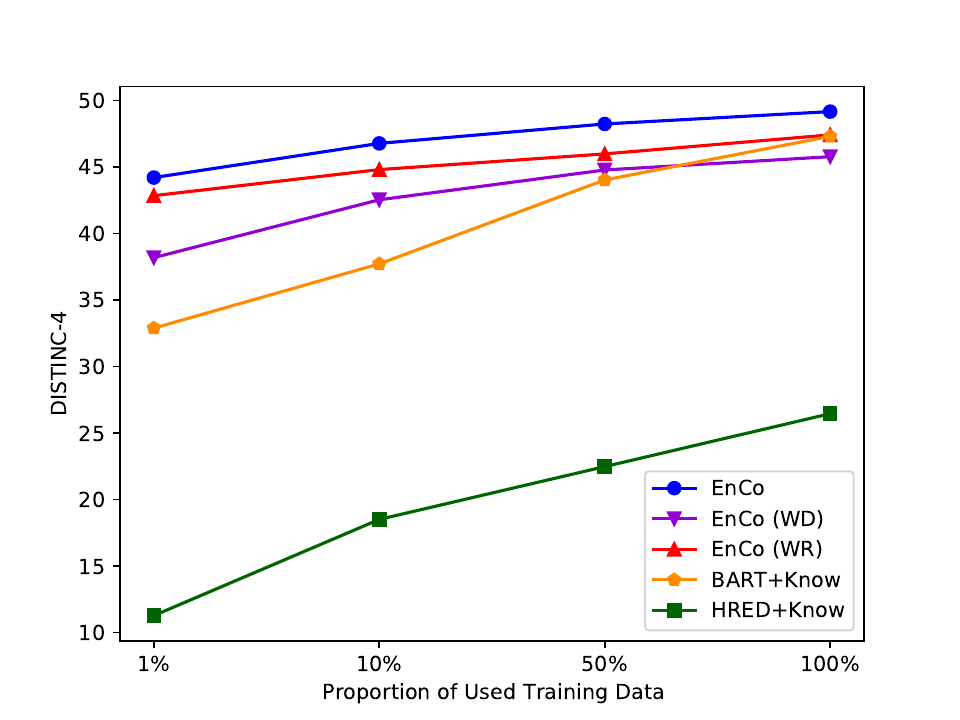}
}
\caption{Few-shot results on DuConv.}
\label{fig:few_shot_results}
\end{figure}

\subsection{Few-Shot Results}

Since the positive samples in EnCo could also be regarded as a data augmentation approach.
Following~\citet{Poddar2022DialAugMU}, we also attempt the following two trivial data augmentation methods to create positive samples: (1) \emph{word deletion} randomly selects 70\% of words in a dialogue context, and replace them with a special token $\texttt{[DEL]}$; (2) \emph{word reordering} randomly samples several pairs of words in a dialogue context (about 30\% of context words), and switch them pairwise. We conduct experiments on DuConv using 1\%, 10\%, 50\% and 100\% training samples, respectively. 

As shown in Figure~\ref{fig:few_shot_results}, EnCo significantly surpasses all comparison models under each setting.
Particularly, under the 1\% setting, our model still achieves the best performances, indicating that our model works well in the few-shot setting as well.
Besides, EnCo (WD) and EnCo (WR) denote using word deletion and word reordering to create positive samples in our EnCo framework, respectively.
The vanilla EnCo outperforms these two variants, indicating the effectiveness of our entity-guided paraphrasing.

\begin{figure}[t]
\centering
\subfigure[Fluency]{
  \includegraphics[width=0.30\linewidth]{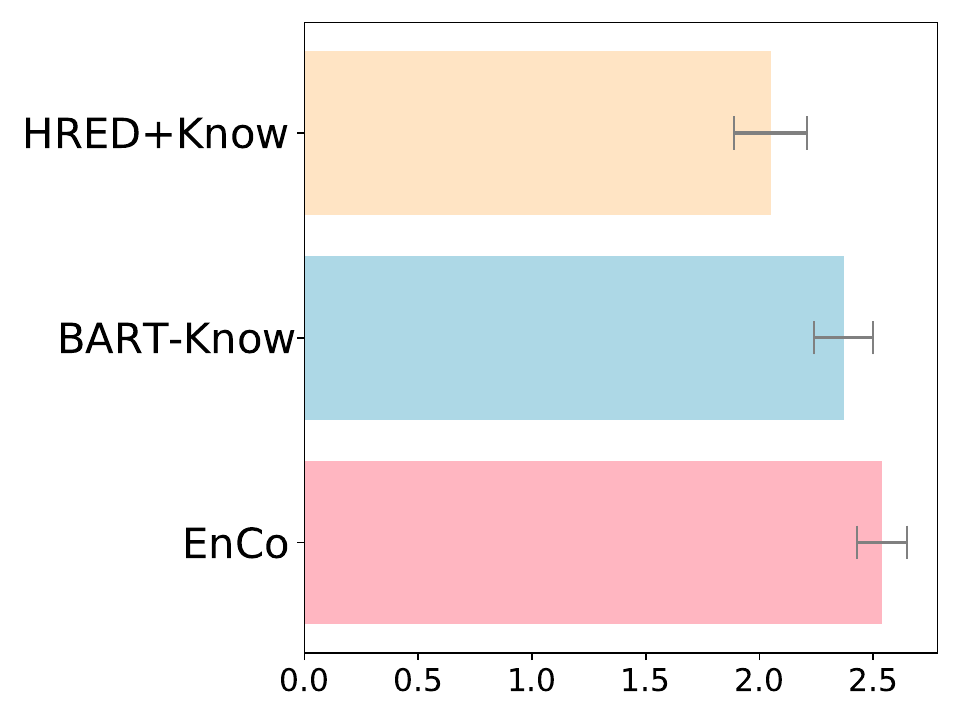}
}
\subfigure[Coherence]{
  \includegraphics[width=0.30\linewidth]{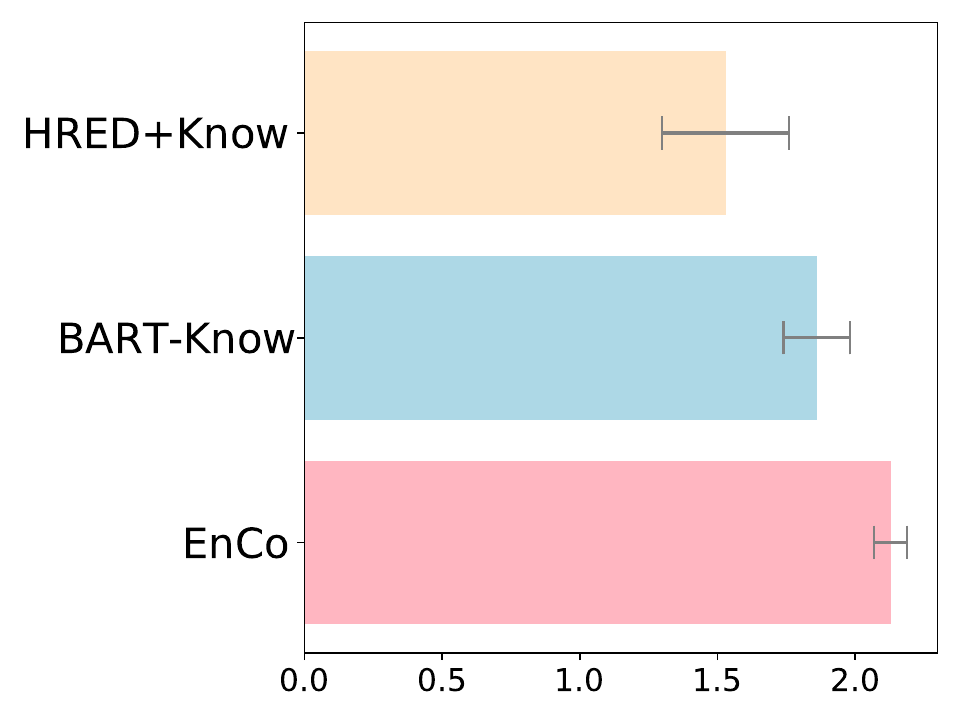}
}
\subfigure[Informativeness]{
  \includegraphics[width=0.30\linewidth]{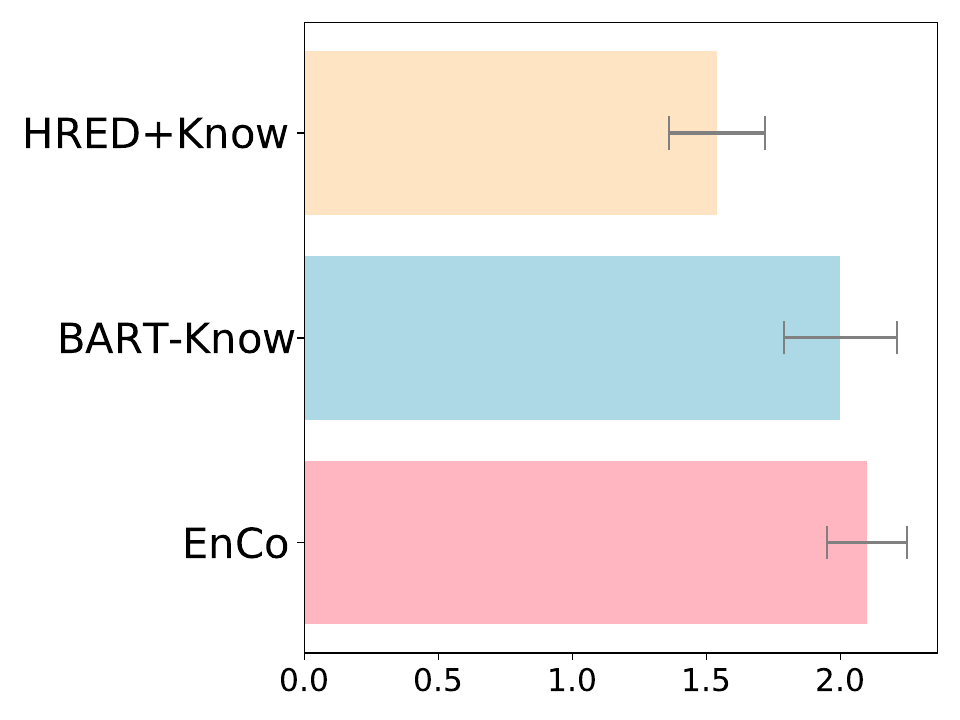}
}
\caption{Results on human study.}
\label{fig:human_results}
\end{figure}

\subsection{Human Study}
Furthermore, we conduct human study on 200 random samples from DuConv. We compare the responses generated by EnCo with the responses generated by HRED+Know and BART+Know.
Five master students are recruited to score the responses in terms of fluency, coherence and informativeness with a 3-point scale.

Figure~\ref{fig:human_results} shows the human study results. EnCo outperforms HRED+Know and BART+Know on all three aspects, which verifies that our method performs better in generating fluent, coherent and informative responses.

\section{Conclusion}
In this paper, we first study the robustness of knowledge-grounded dialogue (KGD) models when faced with real-world noises. We propose an entity-based contrastive learning (EnCo) framework to create positive and negative samples under the guidance of entity information. Then, we use contrastive learning to let the KGD model be aware of semantic-irrelevant and semantic-relevant perturbations. Experimental results on three public benchmark datasets show that our method outperforms state-of-the-art baselines. The robustness study and few-shot study further indicate the superiority of our method under the various types of noises and the few-shot situations, respectively.

\section{Acknowledgments}
This work is supported by the National Natural Science Foundation of China (Grant No. 62102276, 62272334, 62072323), Shanghai Science and Technology Innovation Action Plan (No. 22511104700), the Zhejiang Lab Open Research Project (NO. K2022NB0AB04), the Natural Science Foundation of Jiangsu Province (Grant No. BK20210705, BK20211307), China Postdoctoral Science Foundation (Grant No. 2023M732563), the Natural Science Foundation of Educational Commission of Jiangsu Province, China (Grant No. 21KJD520005), Key Projects of Industrial Foresight and Key Core Technology Research and Development in Suzhou (SYC2022009), Engineering Lab of Bigdata and Intelligence of Jiangsu Province and Project Funded by the Priority Academic Program Development of Jiangsu Higher Education Institutions.

\bibliography{aaai24}

\end{document}